\documentclass[10pt,twocolumn,letterpaper]{article}

\usepackage{cvpr}
\usepackage{microtype}
\usepackage{graphicx}
\usepackage{subfigure}
\usepackage{booktabs} % for professional tables
\usepackage{times}
\usepackage{epsfig}
\usepackage{amsmath}
\usepackage{amssymb}
\usepackage[algo2e]{algorithm2e}
\DeclareMathOperator*{\argmax}{argmax}   % Jhttps://www.overleaf.com/project/5bea150ef28c1123449af00fan Hlavacek
\usepackage{amsmath,amssymb}

\usepackage{booktabs}
\usepackage{tabularx}
\usepackage{arydshln}
\usepackage{array}
\usepackage{makecell}
\usepackage{dsfont}
\usepackage{algorithm}
\usepackage{multicol}
\usepackage[normalem]{ulem}
\usepackage[pagebackref=true,breaklinks=true,letterpaper=true,colorlinks,bookmarks=false]{hyperref}

\cvprfinalcopy % *** Uncomment this line for the final submission

\def\cvprPaperID{****} % *** Enter the CVPR Paper ID here

\ifcvprfinal\fi

\makeatletter
\newcommand*\titleheader[1]{\gdef\@titleheader{#1}}
\AtBeginDocument{%
  \let\st@red@title\@title%
  \def\@title{%
    \vskip-4.0em
    \bgroup\normalfont\large\centering\@titleheader\par\egroup
    \vskip2.0em\st@red@title}
}
\titleheader{\small{Accepted to CVPR 2019 Workshop on Adversarial Machine Learning in Real-World Computer Vision Systems}}

\title{Augmenting Model Robustness with Transformation-Invariant Attacks}
\author{Houpu Yao, Zhe Wang, Guangyu Nie, Yassine Mazboudi, Yezhou Yang, Yi Ren\\
Arizona State University\\
Tempe, Arizona\\
{\tt\small \{hope-yao,zwang559,gnie1,ymazboud,yz.yang,yiren\}@asu.edu}
}

\begin{document}

%%%%%%%%% TITLE
% \title{\LaTeX\ Author Guidelines for CVPR Proceedings}
% \title{Augmenting Model Robustness with Transformation-Invariant Attacks}
% \author{Houpu Yao, Zhe Wang, Guangyu Nie, Yassine Mazboudi, Yezhou Yang, Yi Ren\\
% Arizona State University\\
% Tempe, Arizona\\
% {\tt\small \{hope-yao,zwang559,gnie1,ymazboud,yz.yang,yiren\}@asu.edu}
% }

\maketitle
%\thispagestyle{empty}

% \iffalse
% \iftrue % use space-saving macro
        \newcommand{\cutsectionup}{\vspace*{-0.1in}}
        \newcommand{\cutsectiondown}{\vspace*{-0.07in}}

        \newcommand{\cutsubsectionup}{\vspace*{-0.09in}}
        \newcommand{\cutsubsectiondown}{\vspace*{-0.06in}}

        \newcommand{\cutsubsubsectionup}{\vspace*{-0.17in}}
        \newcommand{\cutsubsubsectiondown}{\vspace*{-0.07in}}

        \newcommand{\cutparagraphup}{\vspace*{-0.12in}}
        \newcommand{\cutparagraphdown}{\vspace*{-0.03in}}

        \newcommand{\cutcaptionup}{\vspace*{-0.1in}}
        \newcommand{\cutcaptiondown}{\vspace*{-0.2in}}

        \newcommand{\cuttablecaptionup}{\vspace*{-0.1in}}
        \newcommand{\cuttablecaptiondown}{\vspace*{-0.15in}}

        \newcommand{\cutequationup}{\vspace*{-0.08in}}
        \newcommand{\cutequationdown}{\vspace*{-0.08in}}

        \newcommand{\cuttableup}{}
        \newcommand{\cuttabledown}{}

        \newcommand{\cut}{{\vspace*{-0.02in}}}
        \newcommand{\cutmore}{{\vspace*{-0.06in}}}
        \newcommand{\negcut}{}
        
        \newcommand{\cutitemup}{{\vspace*{-0.09in}}}
        \newcommand{\cutitemdown}{\vspace*{-0.06in}}
% \else % do not use space-saving macro
        % \newcommand{\cutsectionup}{}
        % \newcommand{\cutsectiondown}{}

        % \newcommand{\cutsubsectionup}{}
        % \newcommand{\cutsubsectiondown}{}

        % \newcommand{\cutsubsubsectionup}{}
        % \newcommand{\cutsubsubsectiondown}{}

        % \newcommand{\cutparagraphup}{}
        % \newcommand{\cutparagraphdown}{}

        % \newcommand{\cutcaptionup}{}
        % \newcommand{\cutcaptiondown}{}

        % \newcommand{\cutequationup}{}
        % \newcommand{\cutequationdown}{}

        % \newcommand{\cuttableup}{}
        % \newcommand{\cuttabledown}{}

        % \newcommand{\cut}{}
        % \newcommand{\cutmore}{}
        % \newcommand{\negcut}{}

%%%%%%%%% ABSTRACT
\begin{abstract}
The vulnerability of neural networks under adversarial attacks has raised serious concerns and motivated extensive research. 
It has been shown that both neural networks and adversarial attacks against them can be sensitive to input transformations such as linear translation and rotation, and that human vision, which is robust against adversarial attacks, is invariant to natural input transformations. 
Based on these, this paper tests the hypothesis that
model robustness can be further improved when it is adversarially trained against transformed attacks and transformation-invariant attacks.
Experiments on MNIST, CIFAR-10, and restricted ImageNet show that while transformations of attacks alone do not affect robustness, transformation-invariant attacks can improve model robustness by 2.5\% on MNIST, 3.7\% on CIFAR-10, and 1.1\% on restricted ImageNet. We discuss the intuition behind this phenomenon.
\end{abstract}

%%%%%%%%% BODY TEXT
\cutsectionup
\cutsectionup
\section{Introduction}
While deep neural networks achieved near-human performance on various machine perception tasks, it is found that these models can be very sensitive to small but carefully designed input perturbations~\cite{szegedy2013intriguing,Goodfellow2014,akhtar2018threat}, thus allowing the attackers to fool a machine in targeted ways by reverse engineering the inputs. Recent studies have demonstrated potential risks in applying neural networks for classification~\cite{nguyen2015deep,moosavi2016deepfool,Kurakin2016,eykholt2018robust}, detection and segmentation~\cite{hendrik2017adversarial_perdestrain,xie2017adversarial_segmentation_detection}, image retrieval~\cite{sharif2016retrieval}, and
reinforcement learning~\cite{Huang2017,kos2017delving}.
Furthermore, it has also been demonstrated that these attacks are successful under real-world settings~\cite{kurakin2016a_adversarial,papernot2017practical_black_box,athalye2017robust_3d_adversarial,evtimov2017robust_physical_attack}, posing much threat to safety-critical applications.

Existing work has revealed that adversarial attacks in the form of small input perturbations can be ineffective under natural input transformations, and therefore random transformations can be used to pre-process the inputs and improve model robustness~\cite{guo2017input_transform}.
It was later shown, however, that such pre-processing essentially creates a gradient obfuscation effect and can be broken by transformation-invariant (robust) attacks ~\cite{athalye2018obfuscated,dong2019evading}. The feasibility of transformation-invariant attacks in real-world applications has also been demonstrated ~\cite{athalye2017robust_3d_adversarial}. Several successful defense have been proposed that do not rely on gradient obfuscation. Notable ones include adversarial training~\cite{tramer2017ensemble,madry2017robust_optimization_PGD} and TRADES~\cite{zhang2019theoretically}. The former directly performs model training through adversarial instead of benign examples; the latter uses adversarial examples to push the decision boundary away from the data distribution.
However, to our best knowledge, no existing approaches explicitly tackle transformation-invariant attacks.

% \cutparagraphup
\begin{figure} [h!]
    \centering
    \includegraphics[width=1.0\linewidth]{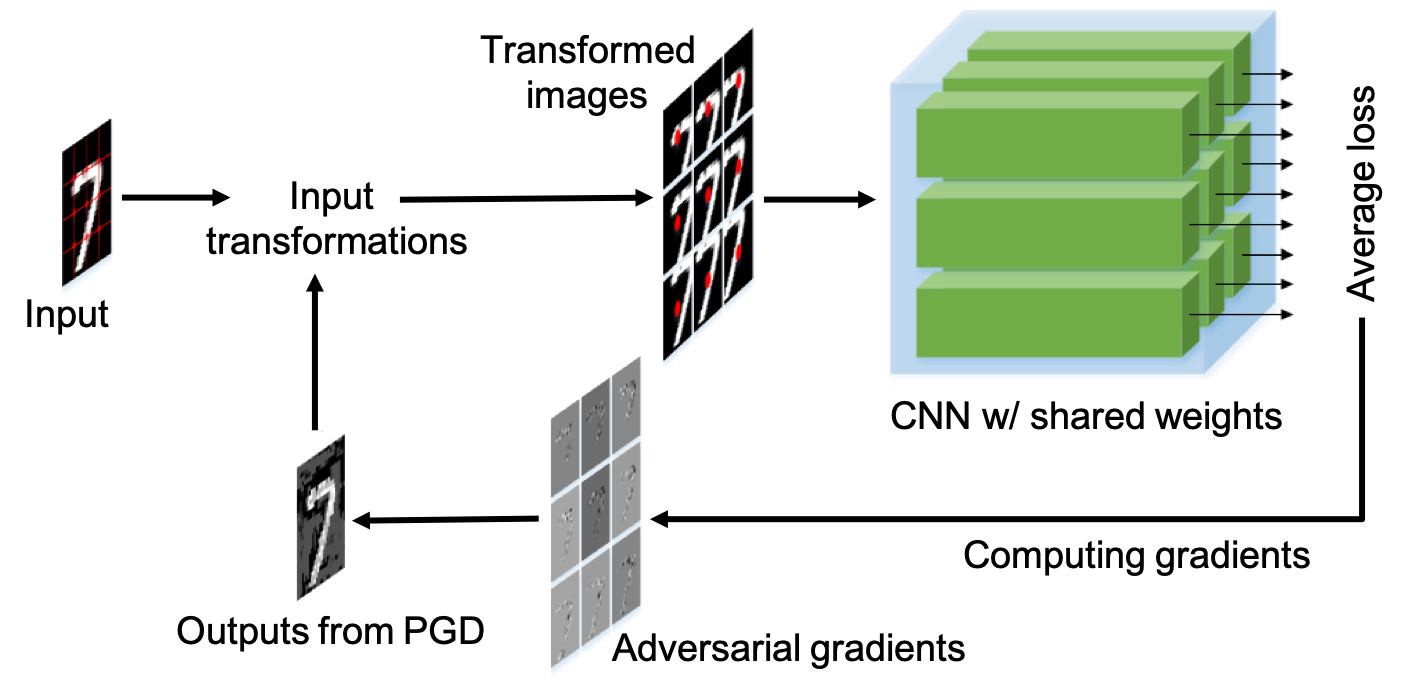}
    \cutcaptionup
    \caption{Schematic of the proposed learning architecture.}
    % , take crop transformation as an instance. (a) Pipeline for the proposed adversarial training method. (b) Locations of crops for tested cropping sizes. Black dots denote the center of the crops.
    \label{fig:scheme}
\end{figure}

\begin{figure}[h]
    \centering
    \includegraphics[width = 1.0\linewidth]{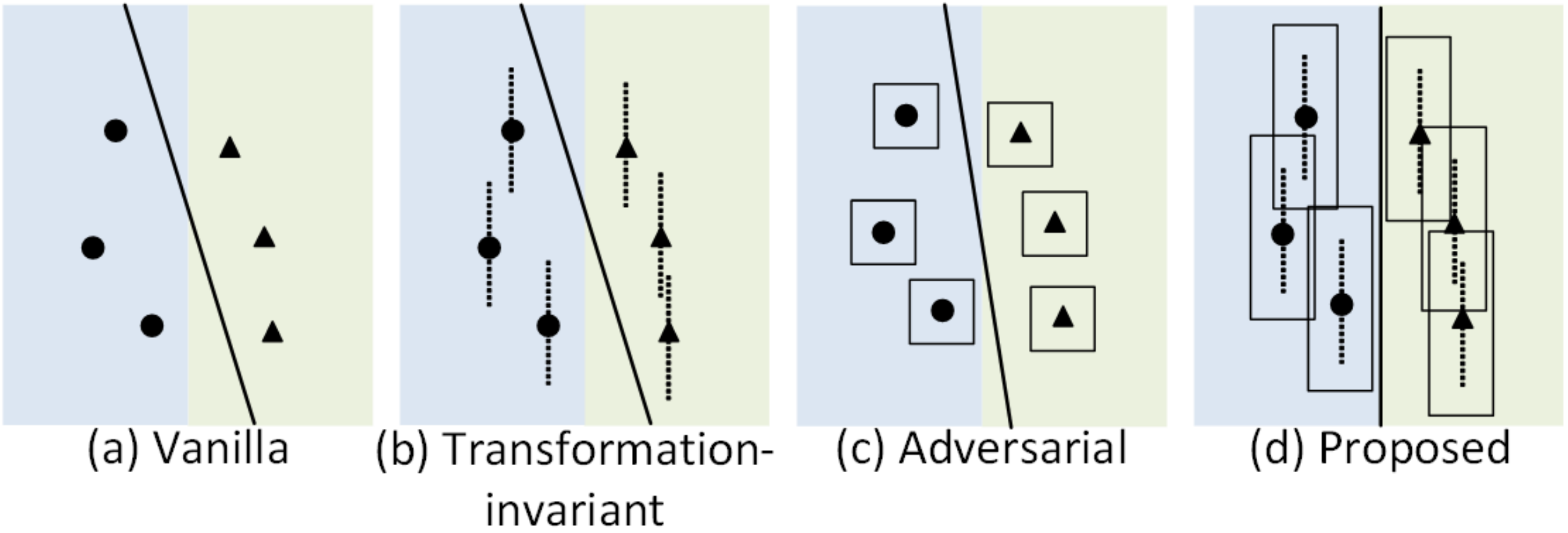}
    \cutcaptionup
    \caption{Comparison between different training methods: (a) Vanilla, (b) transformation-invariant: constraining the classifier with transformations of inputs, by assuming that transformations preserve labels, (c) adversarial training and TRADES: constraining the classifier with $\epsilon$-balls around data points, by assuming that labels are preserved within these balls; (d) proposed: constraining the classifier with both transformations and the transformed $\epsilon$-balls.} 
    \label{fig:diff_method}
    \cutcaptionup
\end{figure}

% \cutparagraphup
This workshop paper is thus motivated to investigate the following hypothesis: \textit{Model robustness can be enhanced through training against transformation-invariant attacks.} 
Specifically, we position our investigation in the context of image classification, and use a classifier that passes an ensemble of input transformations through shared copies of a convolutional neural network, and  aggregates network outputs for decision making (Fig.~\ref{fig:scheme}). As a preliminary study and for implementation feasibility, we consider linear transformations including input cropping, rotation, and zooming. We assume that under appropriate parameter settings, these transformations are content preserving, i.e., the transformed images keep salient features, and can still be correctly classified by human beings. We apply adversarial training to the proposed model, and simulate transformation-invariant adversaries during the attack phases of the training. A comparison between the proposed method and the standard adversarial training is illustrated in Fig.~\ref{fig:diff_method}.

We conducted exhaustive experiments on MNIST, CIFAR-10, and restricted ImageNet, and compaired robustness performance with Madry's et al~\cite{madry2017robust_optimization_PGD}. Empirical results show that while transformations of attacks alone do not affect robustness, transformation-invariant attacks can improve model robustness by 2.5\% on MNIST, 3.7\% on CIFAR-10, and 1.1\% on restricted ImageNet. While a rigorous comparison with TRADES is yet to be performed (to make sure all parametric settings are equal), our current result is on par with those reported in \cite{zhang2019theoretically}, although our method is originated from a different perspective. In addition, we perform model inspection to show that our method does not introduce gradient obfuscation.

While a full analysis is deferred to a full paper, here we provide insights on the connection between this work, which is a direct extension of adversarial training, and TRADES (Fig.~\ref{fig:diff_method}(c-d)). In a binary classification setting, both adversarial training and TRADES seek decision boundaries that avoid cutting through neighborhoods of data points as a result of minimizing the robust error. Since 0-1 loss leads to NP-hard optimization, both utilize surrogate losses for computational feasibility. Between the two, TRADES optimizes a tighter bound of the robust error. However, it should be noted that the definition of robust error relies on the definition of label-preserving operators on the input distribution (e.g., the $\epsilon$-ball). In this work, we essentially impose a more restrictive definition of the robust error by utilizing the fact that salient image features should not be affected by linear transformations~\cite{Ullman2016pnas}. We hypothesize that this constraint can be used in parallel to TRADES, although testing of this hypothesis will be left for future work.   

% At the same time, it was shown that human vision, which is robust to adversaries (including the transformation-invariant ones), is invariant to mild natural input transformations~\cite{Ullman2016pnas}. 

% inspires the question of whether the use of transformation-invariant attacks in adversarial training~\cite{madry2017robust_optimization_PGD} will further , which has so far achieved the state-of-the-art model robustness. 

% which has been proven as one of the few reliable defense mechanisms that do not employ gradient obfuscation~\cite{athalye2018obfuscated}.

% On the other hand, experiments show that adversarial training with an ensemble of different models can be used to effectively avoid gradient obfuscation ~\cite{tramer2017ensemble}. 

% Drawing on all these findings, it is reasonable to question whether adversarial training from an ensemble of models with different natural input transformations may force image classifiers to acquire better robustness.

%%%%%%%%%%%%%%%%%%%%%%%%%%%%%%%%%%%%%%%%%%%%%%%%%%%%%%%%%%%%%%%%%%%%%%%%%%%%%%%%%%%%%%%%%%%%%%%%%%%
% \cutsectionup
\section{Proposed Method}
% \cutsubsectionup
\subsection{Preliminaries}
\cutsubsectiondown
\label{sec:preliminaries}
% \paragraph{Adversarial attacks} 
Given $(x, y) \in \mathcal{D}$ as an image-label pair from dataset $\mathcal{D}$, a classifier $f(\cdot,\theta): \mathbb{R}^d \rightarrow [0,1]^k$ with parameters $\theta$ that maps images to softmax outputs, and a loss function $L(\cdot, \cdot): [0,1]^k \times [0,1]^k \rightarrow \mathbb{R}$, an untargeted attack can be formulated as the problem of finding $x^{adv} = \argmax_{x \in \mathcal{N}_{\epsilon}(x)} L(y, f(x,\theta))$,
where $\mathcal{N}_{\epsilon}(x)$ is a $l_p$ ball around $x$ with radius $\epsilon$. Attacks with $p=\infty$~\cite{szegedy2013intriguing,kurakin2016a_adversarial,kurakin2016b_adversarial,madry2017robust_optimization_PGD}, $p=2$~\cite{carlini2017cw,moosavi2016deepfool}, $p=1$~\cite{chen2017ead}, and $p=0$~\cite{onepixelattack} have been proposed. In this paper we focus on $l_{\infty}$ attack, which is the most widely studied case. Common $l_{\infty}$ attacks include Fast Gradient Sign Method (FGSM), Basic Iterative Method (BIM)~\cite{szegedy2013intriguing}, and Projected Gradient Descent (PGD)~\cite{madry2017robust_optimization_PGD}.

% Various attempts have been made to defend adversarial attacks. Some of the most popular approaches are:
% (1) To suppress the adversarial perturbation of the input. Examples include feature squeezing~\cite{xu2017feature_squeezing}, JPEG compression~\cite{das2018shield}, denoising autoencoder~\cite{gu2014contractive_autoencoder} and its variants ~\cite{liao2018HGD}. 
% (2) To hide useful gradient information from the attacker. Examples include thermometer encoding~\cite{roy2018thermometer_Encoding}, defense distillation~\cite{papernot2016defense_distillation}, and random input transformation~\cite{guo2017input_transform}.
% (3) Adversarial training or robust optimization\cite{kurakin2016a_adversarial,tramer2017ensemble,madry2017robust_optimization_PGD}.

% However, purposefully or not, many existing defense methods from (1) and (2) are based on gradient obfuscation and the robustness is not reliable~\cite{athalye2018obfuscated}. As a result, these models is either vulnerable to black-box attacks~\cite{papernot2016transferability,tramer2017ensemble} or easily defeated by tailored white-box attacks~\cite{athalye2018obfuscated}.

Among all defense mechanisms, adversarial training remains as one of the few that do not suffer from gradient obfuscation~\cite{athalye2018obfuscated}. It can be formulated as a min-max problem:
\begin{equation}
\label{eq:robust_optimization}
    \min_{\theta} \mathop{\mathbb{E}}_{x,y \sim \mathcal{D}}\left[ \max_{x{'}\in \mathcal{N}_{\epsilon}(x)} \displaystyle L (y, f(x{'},\theta)) \right]
\end{equation}
While it is hard to solve Eq.~\eqref{eq:robust_optimization} directly due to the nonconcavity of the inner (attack) problem and the limited capacity of $f$, a common practice is to iteratively find adversaries and train the network on these adversaries~\cite{kurakin2016a_adversarial,madry2017robust_optimization_PGD}. 

% It has been shown that adversarial attacks are transferable among models with different architectures and even learning algorithms~\cite{papernot2016transferability}, suggesting that a surrogate model can be used to perform adversarial attacks on black-box models. 

\vspace{-0.13cm}
\subsection{Transformation-Invariant adversarial training}
\cutsectiondown
\label{sec:proposed}
The proposed network architecture passes an ensemble of transformations of an input through copies of a shared network, before aggregating the network outputs.
Let the shared network be $f(\cdot,\theta)$, and the set of input transformations be $\mathcal{T}$. The final prediction of an input $x$ follows:
\begin{equation}
\label{eq:inference}
y_{pred} = \displaystyle \frac{1}{|\mathcal{T}|}\sum_{T \in \mathcal{T}} f(T(x),\theta).
\end{equation}
And the aggregated loss is defined as:
\begin{equation}
\label{eq:nat_loss}
    J(x,y;\theta) =  \sum_{T \in \mathcal{T}} L(y, f(T(x),\theta))
\cutequationdown
\end{equation}
For classification tasks $L(\cdot,\cdot)$ is the cross-entropy, and standard training is done by solving $\min_{\theta} \mathbb{E}_{x,y\sim \mathcal{D}} \left[J(x,y;\theta)\right]$. 
% The input transformations we tested include cropping, rotation, and zooming. 
% For cropping, we drop out connections from the cropped pixels to the next layer.  
% For our model, we use 9 crops and a cropping size of 20 for MNIST, and 8 crops and a cropping size of 28 for CIFAR-10. The locations of the crops are shown in Fig.~\ref{fig:scheme}b. 
% The number of crops in the case of CIFAR-10 is limited by the fact that we only have 8 GPUs in parallel, each of which handles one copy of the shared network.

% It should be noted that an alternative loss is $L(y, y_{pred})$. We use Eq.~\eqref{eq:nat_loss} to avoid numerical issues in computing the log of sum of exponential terms. Our experiments showed that model robustness is not sensitive to the choice between these two loss definitions. 

Similar to \cite{madry2017robust_optimization_PGD}, we perform adversarial training through: 
\begin{equation}
\min_\theta \mathop{\mathbb{E}}_{x,y \sim D} \left[\max_{x' \in \mathcal{N}_{\epsilon}(x)} J(x',y;\theta)\right]
\cutequationdown
\label{eq:robust}
\end{equation}
It is worth noting that the inner problem in Eq.~\eqref{eq:robust} is an attack robust against \textit{all transformations}, and is similar to the Expectation of Transformation (EoT) method~\cite{athalye2018obfuscated, athalye2017robust_3d_adversarial}. However, to our best knowledge, EoT type of attacks have not yet been incorporated into adversarial training.
We use PGD attacks \cite{madry2017robust_optimization_PGD} to approximate the inner maximization of Eq.~\eqref{eq:robust}.
The parameters for PGD attack are summarized in Tab.~\ref{tab:adv_training_para}.
% , where $a$ and $t$ denote the step size and number of gradient descent steps in one attack, respectively. 

\begin{table}[h]
\centering
\caption{Hyper-parameters used for adversarial training. $\epsilon$: perturbation bound; $a$: step size; $t$: number of steps}
\label{tab:adv_training_para} 
\begin{tabular}{>{\centering\arraybackslash}p{1.5cm}| >{\centering\arraybackslash}p{1cm}| >{\centering\arraybackslash}p{1cm}| >{\centering\arraybackslash}p{1cm}}
\hline
Dataset & $\epsilon$ & $a$ & $t$   \\ 
\hline
MNIST & 0.3 & 0.01 & 40  \\ 
CIFAR-10 & 8/255 & 2/255 & 7 \\ 
\hline
\end{tabular}
\end{table}

\begin{figure} [h!]
    \centering
    \includegraphics[width=0.9\linewidth]{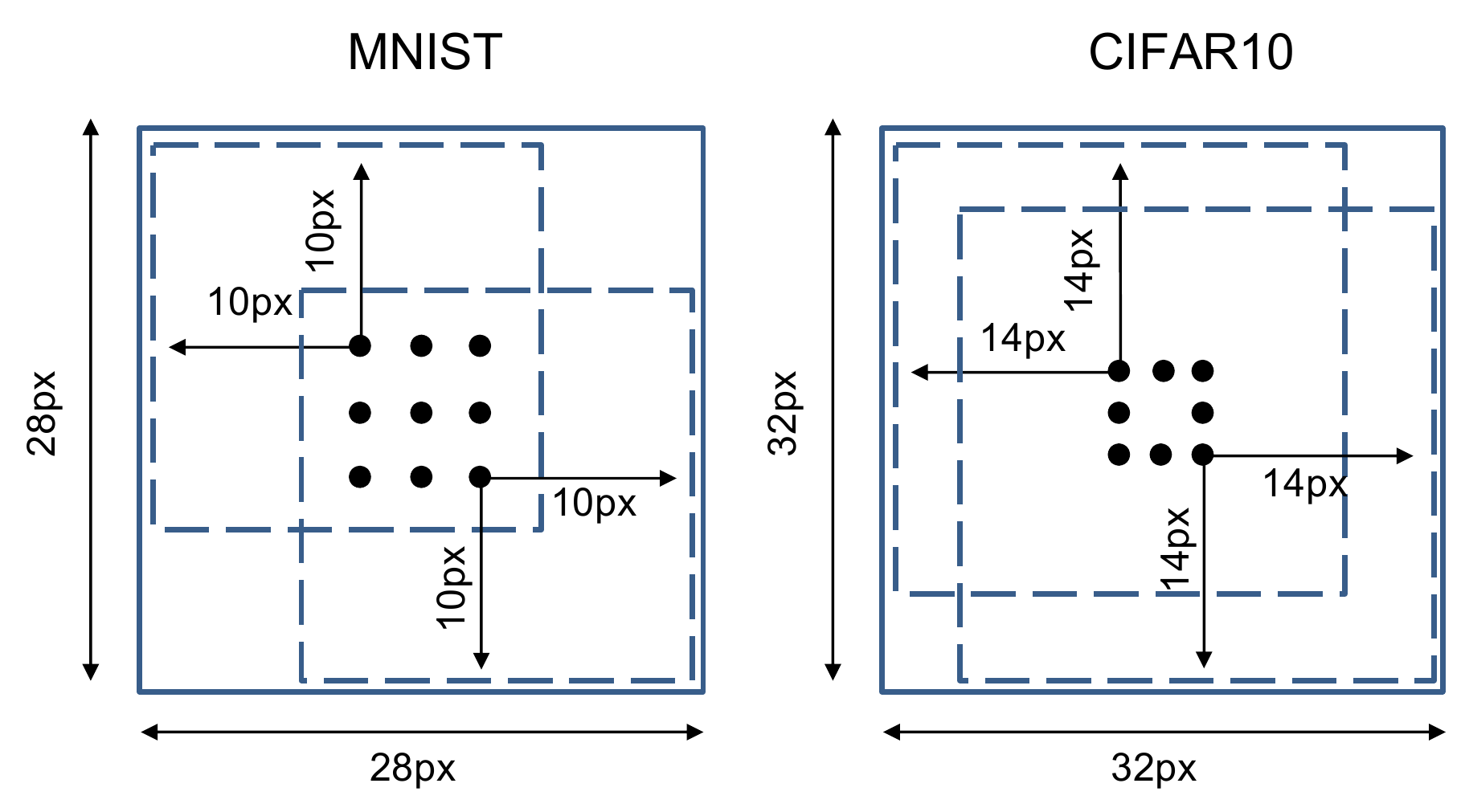}
    \cutcaptionup
    \caption{Locations of crops for tested cropping sizes. Black dots denote the center of the crops.}
    \label{fig:crop_loc}
\end{figure}

Our networks for MNIST and CIFAR-10 follow the wider networks in \cite{madry2017robust_optimization_PGD}. 
We select different cropping operations as the transformation set $\mathcal{T}$ for the two datasets: If not otherwise specified, we use 9 crops each with size 20 for MNIST and 8 crops with size 28 for CIFAR-10 with configurations shown in Fig.\ref{fig:crop_loc}.
We use $B$ or $P$ to denote the baseline network (Madry et al.~\cite{madry2017robust_optimization_PGD}) and the proposed network, respectively, and subscripts $s$ and $r$ to denote standard and adversarial training, respectively. Specifically, $P_r$ represents the proposed model with adversarial training using transformation-invariant attacks.

% We will focus on image cropping as the only transformation applied to the input, and show that with this simple setting our model can already consistently outperform the state of the art on evaluation tasks. Mixed results from applying rotation and zooming will then be discussed. 
% $B_{s}$: the baseline model vanilla, $B_{r}$: the baseline model w/ adv. training, $P_{s}$: Proposed model vanilla, $P_{r}$: Proposed w/ transformation-invariant adv. training
\vspace{-0.6cm}
\section{Results and Discussions}
\cutsectiondown
\label{sec:results}
We evaluate the empirical robustness of the proposed method against various existing attacks with different settings, and perform a thorough investigation on whether our method relies on gradient obfuscation.
We also highlight that training the model with separated attacks for individual input transformations is not effective. Lastly, we show our method is more data efficient than standard adversarial training.

%and its black-box robustness against a variety of source models

\vspace{-0.1cm}
\subsection{Evaluation of Empirical Robustness}

\vspace{-0.2cm}
\paragraph{White-box robustness}

% We use parameter settings in Tab.\ref{tab:white_box_testing_setting} to generate adversarial attacks on testing data to evaluate the white-box robustness of the proposed network. 
%  \cite{madry2017robust_optimization_PGD}: for all attacks, we choose $\epsilon=0.3$ for MNIST and $\epsilon=8/255$ for CIFAR-10.
The white-box robustness of our model is compared with the baseline model on MNIST and CIFAR-10 under FGSM, BIM, PGD, and C\&W~\cite{carlini2017cw} attacks. 
The testing attack parameters for FGSM, PGD, and BIM attacks follow Tab.\ref{tab:adv_training_para}. For C\&W attacks, we use learning rate 0.2 and 40 steps with a Lagrange multiplier of 1.0. 
The results on robustness and clean accuracy are summarized in Tab.\ref{tab:white_box_accuracy}.
The proposed model ($P_r$) achieves the best white-box robustness under all attacks tested. In particular, the proposed method improves white-box PGD robustness from 93.2\% to 95.7\% on MNIST under 40 steps, and from 49.7\% to 54.4\% on CIFAR-10 under 7 PGD steps. 
Furthermore, experiments on Restricted ImageNet shows that our model can improve robustness (clean accuracy) from 92.75\% (96.83\%) of Madry et al. to \textbf{93.85}\% (97.25\%).

\begin{table}[h]
\centering
\caption{Robustness on MNIST and CIFAR-10 against white-box attacks. ``None'': clean test accuracy.}
\begin{tabular}{c||c|c|c|c|c}
\hline
\multicolumn{1}{c||}{~} & \multicolumn{5}{c}{attack}   \\ \hline
MNIST &  None & FGSM & BIM & PGD & C\&W  \\ 
\hline
$B_s$ & 98.9 & 7.0 & 0.0 & 0.0 & 3.2    \\ 
$P_s$(Ours) & \textbf{99.3} & 5.7 & 0.8 & 0.5 & 20.8    \\ 
\hline
$B_r$ & 98.4 & 95.2 & 92.5 & 93.2  & 91.7    \\ 
$P_r$(Ours) & {99.2} & \textbf{96.9} & \textbf{95.0} & \textbf{95.7} & \textbf{96.0}\\
\hline
% \end{tabular}
% \begin{tabular}{c||c|c|c|c|c}
\hline
CIFAR-10 &  None& FGSM & BIM & PGD & C\&W   \\ 
\hline
$B_s$ & 95.2 &  {12.8}   & 0.0 & 4.1 & 0.0  \\ 
$P_s$(Ours) & \textbf{95.6} & 15.2 & 0.0 & 13.0 & 0.3  \\ 
\hline
$B_r$ &  87.3  & 56.4 & 48.36 & 49.7 & 19.51   \\ 
$P_r$(Ours) & {87.9} & \textbf{59.4}  & \textbf{52.9} & \textbf{54.4} & \textbf{22.89} \\
\hline
\end{tabular}
\label{tab:white_box_accuracy}
\vspace{-0.4cm}
\end{table}

\begin{figure}[h]
    \centering
    \includegraphics[width = 1.0\linewidth]{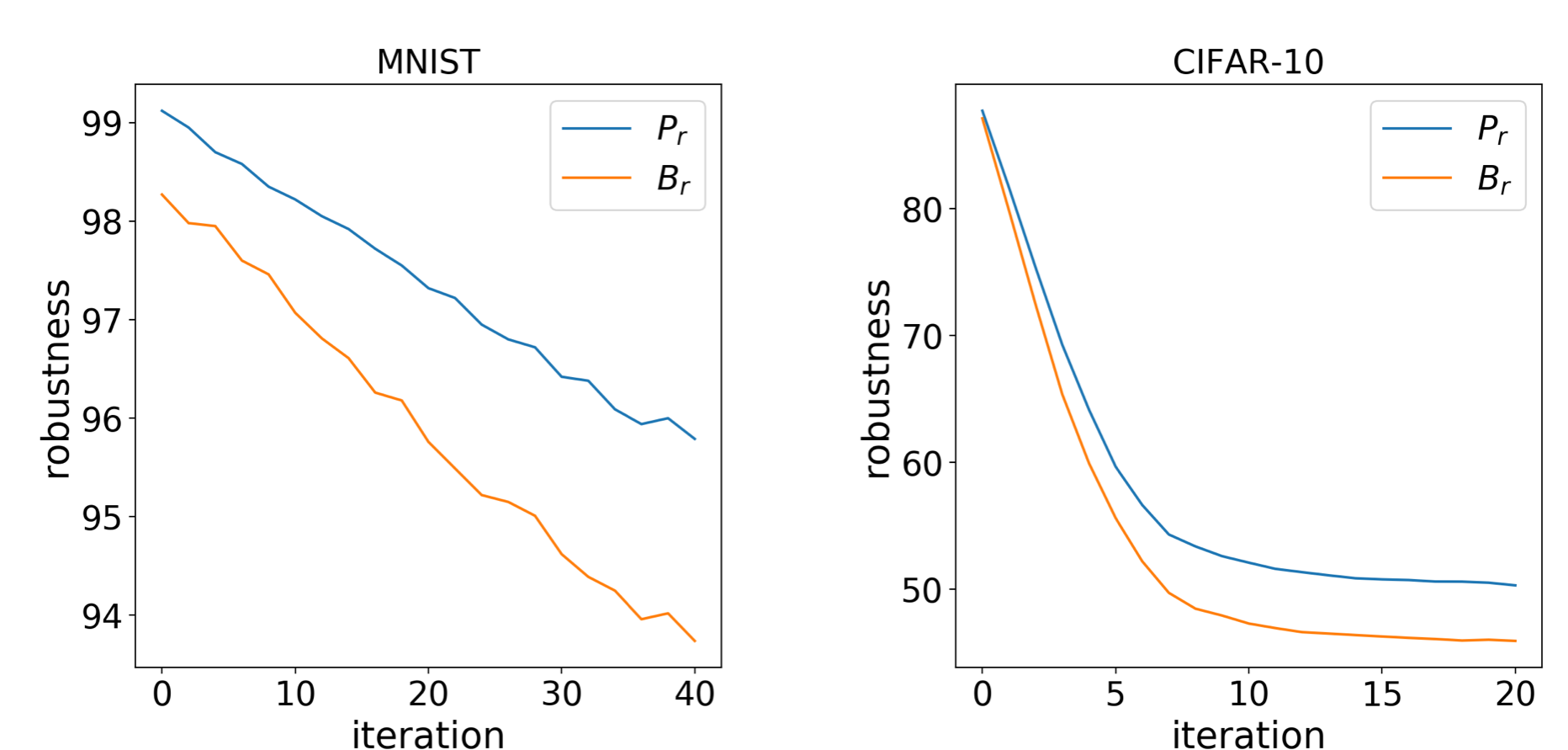}%whitebox_sweep_gain
    \caption{Model robustness for a range of number of PGD steps.} 
    \label{fig:pgd_inc_step_num}
    \vspace{-0.3cm}
\end{figure}

In addition, we test how robustness changes along with the number of iterations ($t$) in PGD attacks.
To do so, we increase $t$ from 0 to 20 while fixing the attack bound $\epsilon$ and the step size $a$, and compare the performance between $P_r$ and $B_r$ (Fig.\ref{fig:pgd_inc_step_num}). The proposed model consistently out-performs the baseline. 
% For CIFAR-10, when we increase number of PGD steps to 20, proposed model still has a robustness of 50.3\% compared with 45.9\% on baseline model. 
It should also be noted that our model is comparable to TRADES~\cite{zhang2019theoretically}, although a more rigorous comparison is needed (e.g., \cite{zhang2019theoretically} uses ResNet-18 for CIFAR-10, while we followed the model in \cite{madry2017robust_optimization_PGD}). Specifically, on CIFAR-10 with 20-step PGD and $\epsilon=8/255$, TRADES achieves robustness (clean accuracy) of $56.6\%$ ($84.9\%$) for $1/\lambda = 6$ and $49.1\%$ ($88.6\%$) for $1/\lambda = 1$. In comparison, our model achieves a robustness (clean accuracy) of $50.3\%$ (87.7\%). 

\begin{figure}[h]
    \centering
    \includegraphics[width = 1.0\linewidth]{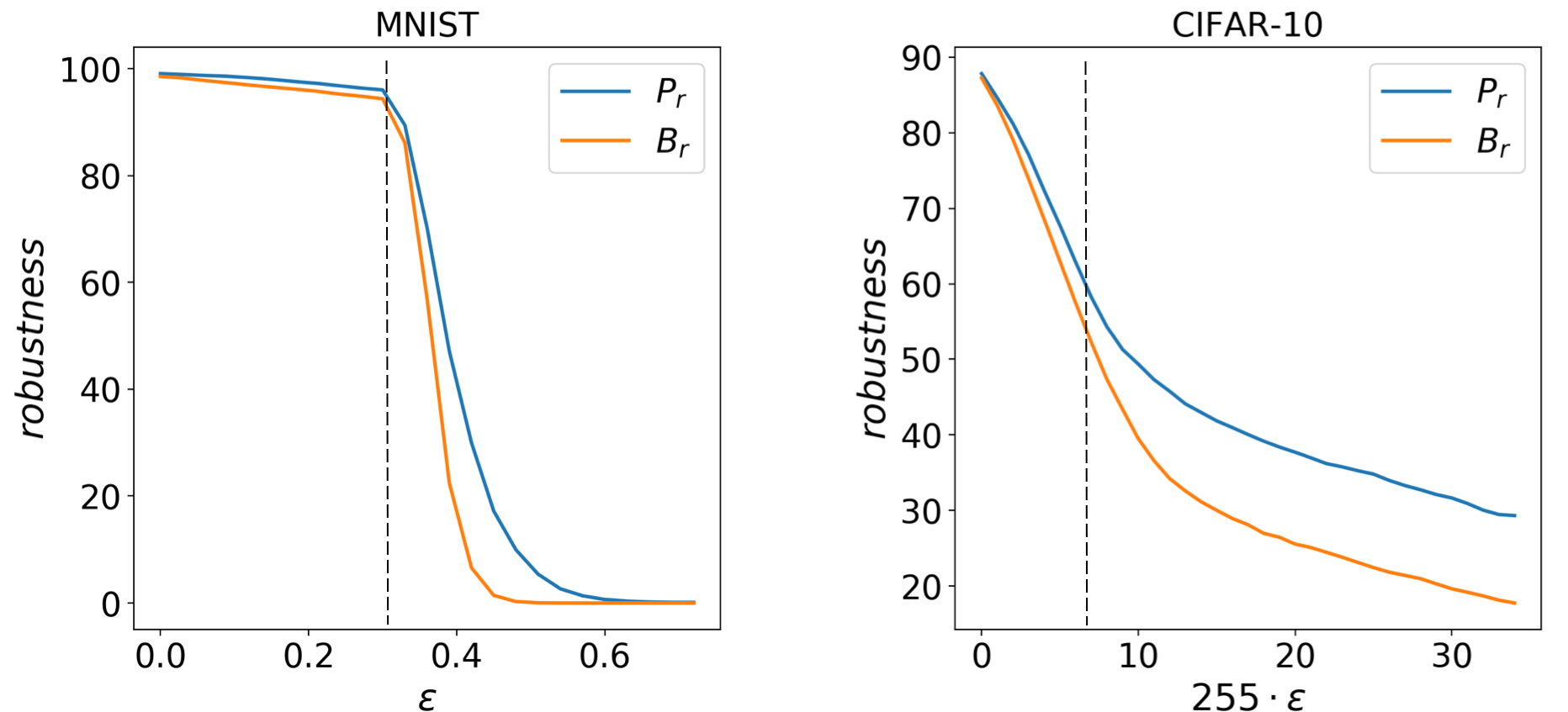}%whitebox_sweep_gain
    \caption{Model robustness for a range of attack bounds. The attack bounds used for training are marked as black vertical lines.} 
    \label{fig:beyond}
\vspace{-0.5cm}
\end{figure}

Furthermore, we compare the white-box robustness under PGD attacks beyond the attack bounds $\epsilon$ used during adversarial training. For MNIST, we test $\epsilon \in [0,1]$ where $\epsilon=1$ represents the maximal $l_{\infty}$ attack strength. For CIFAR-10, we test $\epsilon \in [0, 35/255]$. The comparisons are shown in Fig.~\ref{fig:beyond}. Still, the proposed method exhibits consistently higher robustness than the baseline under all attack bounds.

\cutparagraphup
\paragraph{Different Transformations}
We now test how the proposed method performs using different input transformations. First we study the influence of the number of input crops and the cropping size on model robustness using MNIST: For the former, we fix the cropping size to 20 and vary the number of crops from 1 to 64. For the latter, we fix the number of crops to 9 and vary the cropping size from 12 to 24. The robustness of these models under white-box PGD attacks with settings in Tab.\ref{tab:adv_training_para} are summarized in Tab.~\ref{tab:parametric}.
It can be seen that increasing the number of crops helps to improve both clean and adversarial accuracy, with diminishing effect. On the other hand, a sweet spot exists for the cropping size: Larger cropping sizes tend to improve clean test accuracy, yet inevitably lead to reduced robustness. 

\begin{table}[ht]
\vspace{-0.1cm}
\caption{Parametric study on the number of crops and the cropping size used in the proposed model}
\centering
\label{tab:parametric}
\begin{tabular}{c|ccccc}
\hline
\hline
cropping size & 12 & 16 & 20 & 24 & 28 \\
\hline
clean testing & 11.3 & 98.6 & \textbf{99.2} & 99.1 & 98.4\\ 
PGD white-box & 11.1 & 94.2 & \textbf{95.7} & 94.9 & 93.2\\ 
\hline
\hline
\# of crops & 1 & 4 & 9 & 36 & 64   \\ \hline
clean testing & 98.3 & 99.0 & \textbf{99.2} & \textbf{99.2} & \textbf{99.2} \\ 
PGD white-box & 92.1 & 95.1 & 95.7 & \textbf{96.1} & \textbf{96.1} \\ 
\hline
\end{tabular}
\vspace{-0.2cm}
\end{table}

Another experiment is performed on testing the effect of using input rotation and zooming as transformations. For rotation, we use the built-in rotation function from TensorFlow and equal rotation intervals up to 4 degrees.
For zooming, we first apply cropping and then re-scale the image back to the original size through bi-linear interpolation.
Tab.\ref{tab:comp_transofmration} shows mixed results.
On MNIST, rotation yields more robust models while zooming does not; on CIFAR-10, model robustness improves with mild rotation angles (maximum 4 degrees); yet larger angles, e.g. maximum 30 degrees, show limited effects (not listed here). The reason could be that larger rotations introduces more features that require a larger network capacity to learn.
\begin{table}[h]
\vspace{-0.1cm}
\centering
\caption{Robustness (accuracy) of models trained on transformation-invariant attacks for MNIST and CIFAR-10. The reference on baseline model $B_r$ is also listed on the top of the table.}
\label{tab:comp_transofmration}
\begin{tabular}{c|c|c|c}
\hline
\multicolumn{4}{c}{MNIST,   $B_r$: 93.2 (98.4)}  \\ 
\hline
$|\mathcal{T}|$ & 1 & 4 & 9   \\ 
\hline
cropping  & 92.1 (98.3) & 95.1 (99.0) & 95.7 (99.2)\\ 
rotation & 93.3 (98.7) & 95.5 (99.1) & 96.1(99.3)  \\ %95.4(98.8) 
zooming  & 90.7 (99.2) & 91.5 (98.5) &  93.6 (99.2) \\ 
% \end{tabular}
% \begin{tabular}{c|c|c|c}
\hline
\multicolumn{4}{c}{CIFAR-10,   $B_r$: 47.3 (87.3)}  \\ 
\hline
$|\mathcal{T}|$ & 1 & 4 & 8   \\ 
\hline
cropping & 45.1 (80.5) & 53.3 (83.1) & 54.4 (87.9) \\
rotation & 46.5 (78.6) & 54.4 (86.8)  & 53.8 (87.8) \\ %48.1 (78.9)
\hline
\end{tabular}
\vspace{-0.2cm}
\end{table}

Lastly, we note that combining different transformations can lead to additional improvement in robustness. Specifically, we tested the white-box PGD robustness of the combination of two models on MNIST, trained separately with cropping (cropping size of 20, 9 crops) and rotation (4 orientations, max angle of 4 degrees) as the input transformations. The combined model outputs aggregated decisions from these two models, and reaches a white-box robustness of 97.2\%, while the two individual models have individual robustness of 95.7\% (cropping) and 95.5\% (rotation).

% It is worth noting that combining different transformations may lead to extra improvement in robustness. Specifically, we tested the combination of two models on MNIST, trained separately with cropping (cropping size of 20, 9 crops) and rotation (4 orientations). The model reaches 97.2\% white-box robustness, while the individual models have 95.7\% with cropping and 95.5\% with rotation.
% \cutparagraphup
\subsection{Gradient Obfuscation}
We have shown that our model achieves high white-box robustness. However, as discussed in \cite{athalye2018obfuscated}, models with gradient obfuscation (shattering, masking, and explosion/vanishing) can have close to zero robustness. We first note that the proposed model
does not create shattering since it has no randomness;
and does not create gradient explosion or vanishing since it does not contain long recurrence. 
In the following, we investigate whether gradient masking exists in our model. We use standard tools to this end, including black-box attacks, non-gradient based attacks, and visualization of the loss landscape.
% (3) There are two pieces of evidence show that our model does not create gradient masking: First, the black-box accuracy of our model is higher than its white-box accuracy; second, from the \textbf{loss landscapes} around random test points shown in Fig.~\ref{fig:landscape}, the gradient directions, in comparison with random orthogonal directions, are informative and smooth. Therefore, the reported robustness improvements are reliable.

\cutparagraphup
\paragraph{Black-box robustness}
We perform PGD attacks on four source models: vanilla ($B_{s}$) and robust ($B_{r}$) versions of Baseline model, and those ($P_{s}$ and $P_{r}$) of the proposed. 
We again use the attack settings listed in Tab.\ref{tab:adv_training_para} to generate adversarial samples on the source model. Test results on the target model are summarized in Tab.\ref{tab:blackbox_accuracy}, where rows and columns correspond to different tests and source models, respectively. Diagonal elements in the table corresponds to the white-box robustness. The proposed model achieves good robustness in these tasks. Since gradient masking often results in higher robustness for white-box rather than black-box attacks, this result suggests that the proposed model does not relay on gradient masking.
\begin{table}[h]
\centering
\vspace{-0.1cm}
\caption{Robustness on MNIST/ CIFAR-10 against black-box PGD attacks. Columns are the source model to obtain adversaries, rows are different target models to be attacked. Diagonal terms in the table are white box robustness.}
\begin{tabular}{c||c|c|c|c}
\hline
% \backslashbox{T}{S}
& $B_{s}$& $P_{s}$ & $B_{r}$  & $P_{r}$   \\ 
\hline
\hline
$B_{s}$ & \sout{0.0/0.0}  & 8.8/0.2 & 85.6/79.8 & 94.8/68.0  \\ 
\hline
$P_{s}$& 39.3/ 0.2 & \sout{0.6/ 0.0} & 63.6/ \textbf{81.1} & 89.5/ \textbf{69.7}  \\ 
\hline
$B_{r}$  & 93.2/ \textbf{96.7} & 96.5/ 86.1 & \sout{93.2/ 47.4} & 95.5/ 66.9  \\ 
\hline
$P_{r}$ &\textbf{97.9}/ {86.2} & \textbf{97.4}/ \textbf{86.3}   & \textbf{98.2}/ 69.5  & \sout{\textbf{95.7}/ 54.4} \\
\hline
\end{tabular}
\label{tab:blackbox_accuracy}
\vspace{-0.2cm}
\end{table}

\begin{figure}[h!]
    \centering
    \includegraphics[width = 1.0\linewidth]{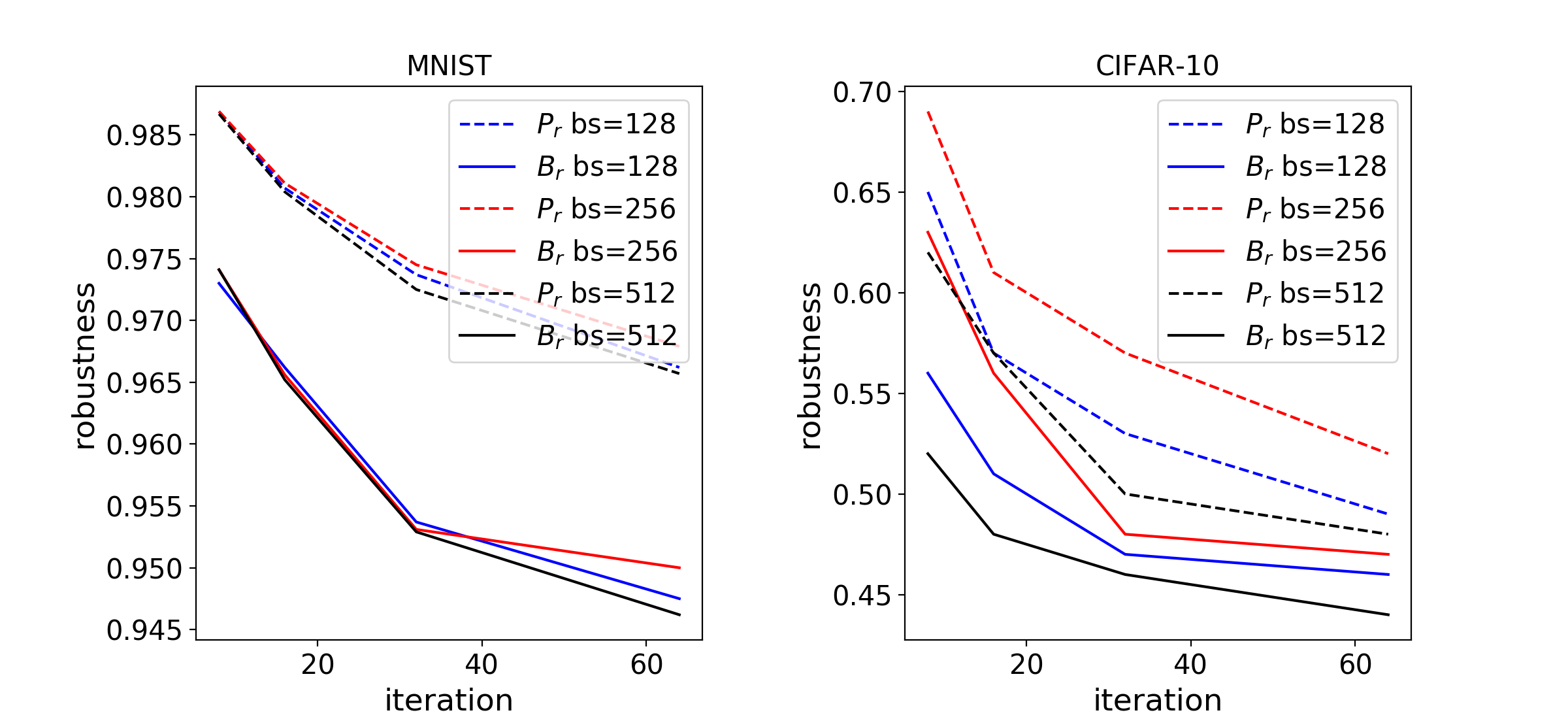}
    \cutcaptionup
    \caption{Network robustness under SPSA attack.}
    %  X axis is the number of iterations, Y axis is the robustness. 
\label{fig:spsa_result}
\vspace{-0.3cm}
\end{figure}

\cutparagraphup
\cutparagraphup
\paragraph{None-gradient based attacks}
We tested the performance of $B_r$ and $P_r$ under white-box SPSA attacks~\cite{uesato2018spsa}. As shown in Fig.\ref{fig:spsa_result}, the proposed method consistently out-performs the baseline under different SPSA max. iteration numbers and batch sizes. Since SPSA uses estimated gradient instead of direct differentiation, it is less likely to be affected by gradient masking. Our model is able to obtain higher robustness on SPSA, which again indicates that the proposed model does not rely on gradient masking.

\begin{figure}[h]
    \centering
    \includegraphics[scale=0.45]{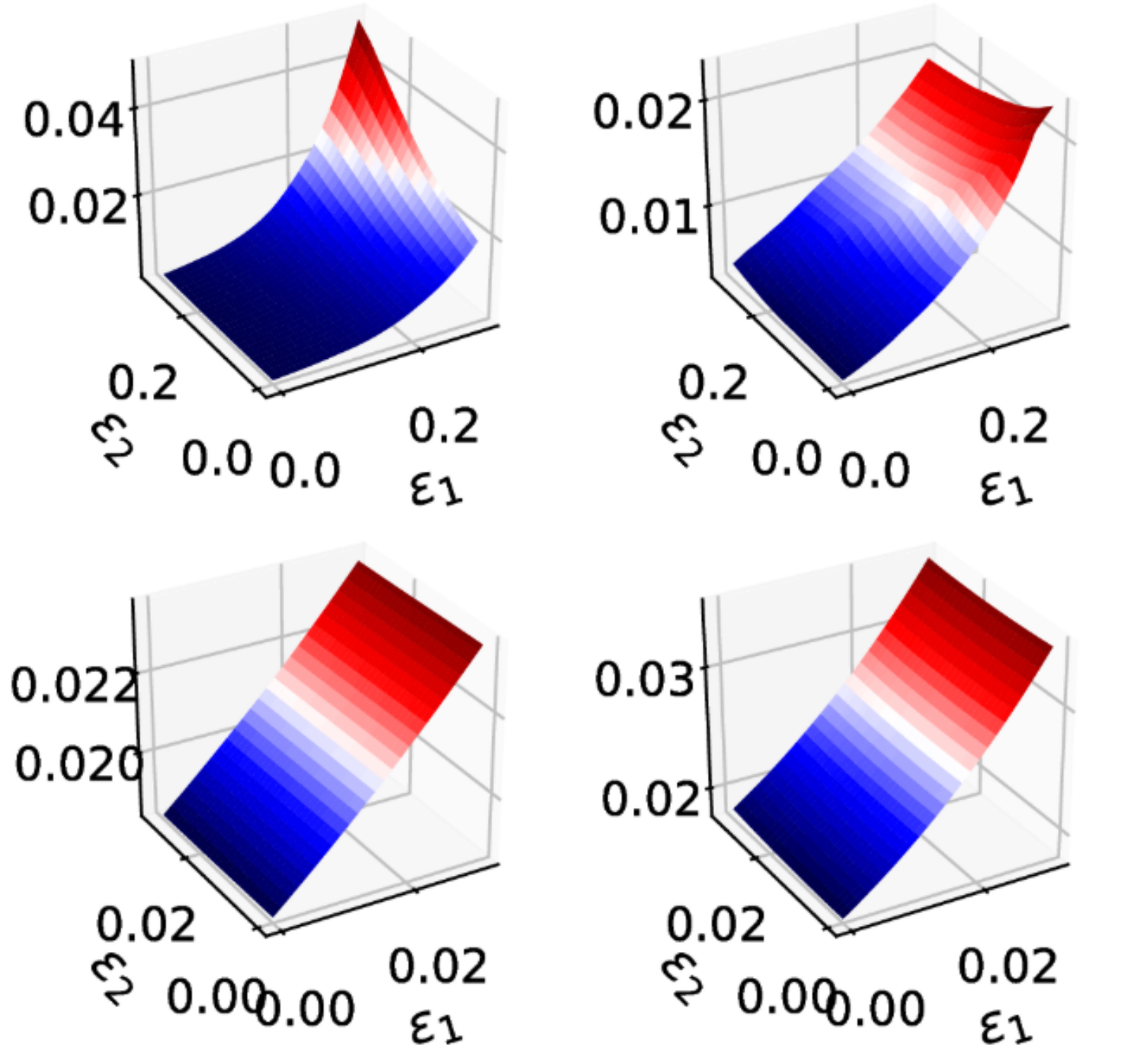}
    % \cutcaptionup
    \caption{Loss landscapes around random test samples on MNIST (top row) and CIFAR-10 (bottom row). $\epsilon_1$ is the adversarial gradient direction and $\epsilon_2$ is a random direction orthogonal to $\epsilon_1$. }
    \label{fig:landscape}
\vspace{-0.4cm}
\end{figure}

\cutparagraphup
\paragraph{Loss landscape}
Lastly, we visualize the loss landscapes of $P_r$ around random test data points. As shown in Fig.~\ref{fig:landscape}, the landscapes for $P_r$ trained on both MNIST and CIFAR-10 are smooth. This further confirms that the proposed model does not rely on gradient masking.

\subsection{Performance Analysis}

\paragraph{Effect of transformation-invariant attacks}
To better understand the effect of transformation ensemble on model robustness, an ablation study is conducted.
% and monitor how the robustness changes in each of the cases. 

{\it No ensemble in the training phase}: We remove the influence of ensemble on the generation of adversaries during training phase, in which case the training becomes standard adversarial training with training data replaced by their cropped copies. For example, when using nine cropping windows, it leads to a 9-fold data augmentation. During adversarial training, this experiment setting results in adversaries that target individual transformations, and are not necessarily transformation invariant. In the test phase, we perform the same ensemble operation as in Eq.~\eqref{eq:inference}. On MNIST, these settings lead to a model robustness of 93.4\% and clean test accuracy of 98.7\%, which is comparable to the baseline model and worse than the proposed. This experiment reveals the critical role of transformation-invariant attacks in improving robustness from standard adversarial training. 
We conjecture that when the model is attacked by adversaries for individual transformations, these adversaries may lead to contradictory gradient directions for model refinement, i.e., refining the model with respect to attacks for one particular transformation may not help (or even worsen) the robustness under other transformations.

\begin{figure}[h]
\vspace*{-0.3cm}
    \centering
    \includegraphics[scale=0.25]{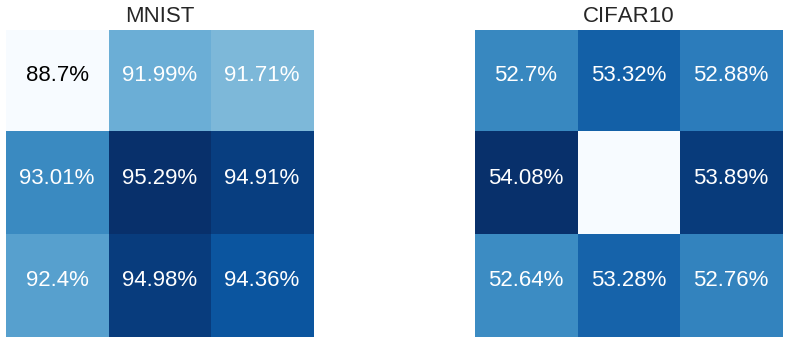}
    \caption{White-box robustness of individual network copies within the ensemble under PGD attacks. Locations in the grid correspond to cropping locations. 
    }
    \label{fig:acc_ablation}
\vspace*{-0.85cm}
\end{figure}

{\it No ensemble in the test phase}: Here we train the model as proposed, and test the white-box robustness of each individual network copy in the ensemble. The copies are only different in their input transformation layers. As shown in Fig.\ref{fig:acc_ablation}, the robustness of individual copies are slightly lower than the robustness of the ensemble (95.7\% and 54.4\%), suggesting that the ensemble also contributes to improvement in robustness.
%  range from 88.7\% to 95.3\% on MNIST, and 52.6\% to 54.1\% on CIFAR-10.

\paragraph{Learning efficiency}
% As discussed in \ref{tsipras2018no_free_lunch}, there exists a trade-off between robustness and accuracy. It is also found that this trade off is depended on the input distribution \cite{ding2018data_distribution}. 
% and requires much larger amount of data \cite{schmidt2018sample_complexity} to alleviate.
% In our case, while using the same network capacity and without changing the data distribution, we are able to perform much better in robust generalization. 
% While the trade-off still holds true for our model, it is found this trade-off is shifted away from the baseline model.
% We train the proposed and the baseline model models with increasing training data sizes and compare model robustness and accuracy along the data size. 
Lastly, we test how $B_r$ and $P_r$ performs when they are trained with different amount of training data. PGD attacks with parameters from Tab.\ref{tab:adv_training_para} are applied in both training and testing. Results are shown in Tab.\ref{tab:dataset_size}. It is obvious that the proposed model consistently requires less training data for the same robustness or accuracy levels. 
% Considering that cropping is a dimension-reducing transformation, the result here is consistent with the analysis from \cite{schmidt2018sample_complexity} that the sample complexity (i.e., the data size) for reaching a certain level of model robustness is related to the input dimensionality.

\begin{table}[h]
\vspace*{-0.2cm}
\centering
\caption{Comparison on learning efficiency. A: clean test accuracy, R: white-box robustness under PGD attacks}
\begin{tabular}{c||c|c|c|c|c|c}
\hline
\multicolumn{1}{c||}{~} & \multicolumn{6}{c}{$|\mathcal{D}|$}  \\ \hline
MNIST  & 0.3k & 1k & 3k & 10k & 30k &  60k \\ 
\hline
A($B_{r}$)  &  91.7 & \textbf{96.6} & 96.5 & 97.8 & 98.5 & 98.4  \\ 
% \hline
A($P_{r}$)&  \textbf{93.4} & 96.3 & \textbf{97.6} & \textbf{98.5} & \textbf{98.8} & \textbf{99.2} \\ 
\hline
R($B_{r}$)  &  35.0 & 73.6 & 80.5 & 86.7 & 88.0 &  93.2 \\ 
% \hline
R($P_{r}$)& \textbf{68.2} & \textbf{82.6} & \textbf{89.0} & \textbf{93.7} & \textbf{95.6} &  \textbf{95.7}  \\
\hline
% \end{tabular}
% \begin{tabular}{c||c|c|c|c|c|c}
\hline
CIFAR-10 &   0.3k & 1k & 3k & 10k & 30k &  60k \\ 
\hline
A($B_{r}$)  & 43.1 & 52.2 & 62.5 & \textbf{73.2} & 80.0 & 87.3 \\ 
% \hline
A($P_{r}$)&   \textbf{43.3} & \textbf{55.0} & \textbf{67.4} & 72.1 & \textbf{80.2} & \textbf{87.9} \\ 
\hline
R($B_{r}$)  &  10 & 12.8 & 21.4 & 33.1 & 45.3 &  47.4 \\ 
% \hline
R($P_{r}$)&  \textbf{12.7} & \textbf{17.6} & \textbf{27.2} & \textbf{37.2} & \textbf{47.8} & \textbf{54.5} \\
\hline
\end{tabular}
\label{tab:dataset_size}
\vspace*{-0.5cm}
\end{table}

\section{Conclusion}
In this paper we investigated a learning architecture that incorporates input transformations into adversarial training, and showed that the resultant model (1) improves empirical robustness over standard adversarial training; (2) is free of gradient obfuscation; and (3) is more data efficient. Importantly, we also showed that while constraining the model to be transformation-invariant (through data augmentation) does not help improve model robustness, incorporating transformation-invariant attacks in training plays a critical role in achieving this goal.
\pagebreak
{\small
\bibliographystyle{ieee}
\bibliography{egbib}
}

\end{document}